\documentclass[conference,hidelinks]{IEEEtran}
\IEEEoverridecommandlockouts
\usepackage{graphicx}
\usepackage{textcomp}
\usepackage{times}
\usepackage{epsfig}
\usepackage{graphicx}
\usepackage{booktabs}
\usepackage{enumitem}
\usepackage{array}
\usepackage{makecell}
\usepackage{hyperref}
\usepackage[table]{xcolor}
\usepackage{cite}
\usepackage{amsmath,amssymb,amsfonts}
\usepackage{algorithmic}
\usepackage{graphicx}
\usepackage{multirow}
\usepackage{authblk}

\usepackage{textcomp}
\usepackage{xcolor}
\def\BibTeX{{\rm B\kern-.05em{\sc i\kern-.025em b}\kern-.08em
    T\kern-.1667em\lower.7ex\hbox{E}\kern-.125emX}}

\title{STGV: Spatio-Temporal Hash Encoding for Gaussian-based Video Representation}
\begin{document}

\newif\ifanonymous
\anonymousfalse  

\ifanonymous
    \author{Anonymous ICME submission}
    \maketitle
\else
    \author{
Jierun Lin$^{1,*}$, Jiacong Chen$^{2,3,*}$, Qingyu Mao$^{1}$, Shuai Liu$^{2,3}$,
Xiandong Meng$^{4}$\\ Fanyang Meng$^{4}$, Yongsheng Liang$^{1,2,3,\dagger}$\\
\small
    $^{1}$College of Electronics and Information Engineering, Shenzhen University\\
$^{2}$College of Applied Technology, Shenzhen University\\
$^{3}$School of Artificial Intelligence, Shenzhen Technology University\\
$^{4}$Research Center of Networks and Communications, Pengcheng Laboratory\\
Email: liangys@szu.edu.cn
}
\renewcommand{\thefootnote}{\fnsymbol{footnote}}
\maketitle
\footnotetext[1]{Equal contribution.}
\footnotetext[2]{Corresponding author}
\fi

\begin{abstract}
2D Gaussian Splatting (2DGS) has recently become a promising paradigm for high-quality video representation.
However, existing methods employ content-agnostic or spatio-temporal feature overlapping embeddings to predict canonical Gaussian primitive deformations, which entangles static and dynamic components in videos and prevents modeling their distinct properties effectively.
This results in inaccurate predictions for spatio-temporal deformations and unsatisfactory representation quality. 
To address these problems, this paper proposes a Spatio-Temporal hash encoding framework for Gaussian-based Video representation (STGV).
By decomposing video features into learnable 2D spatial and 3D temporal hash encodings, STGV effectively facilitates the learning of motion patterns for dynamic components while maintaining background details for static elements.
In addition, we construct a more stable and consistent initial canonical Gaussian representation through a key frame canonical initialization strategy, preventing feature overlapping and a structurally incoherent geometry representation.
Experimental results demonstrate that our method attains better video representation quality (+0.98 PSNR) against other Gaussian-based methods and achieves competitive performance in downstream video tasks.

\end{abstract}


\section{Introduction}
\label{sec:intro}
Video representation is an important task in computer vision and signal processing, which has numerous applications in fields such as video compression~\cite{wiegand2003overview}, inpainting~\cite{kim2019deep} and interpolation~\cite{niklaus2017video}.
The emergence of Implicit Neural Representation (INR) presents a promising approach for video representation~\cite{chen2021nerv}, which utilizes a neural network to learn an implicit continuous mapping from input temporal coordinates to the corresponding frames.
While INRs~\cite{li2022Enerv, chen2023hnerv} are capable of capturing and retaining video details with remarkable performance by employing high-capacity networks, this often results in slow training speed, excessive memory requirements and prolonged decoding times~\cite{liu2025d2gv, zeng2025instant}.

\begin{figure}[t]
\begin{center}  
\includegraphics[width=1.0\linewidth]{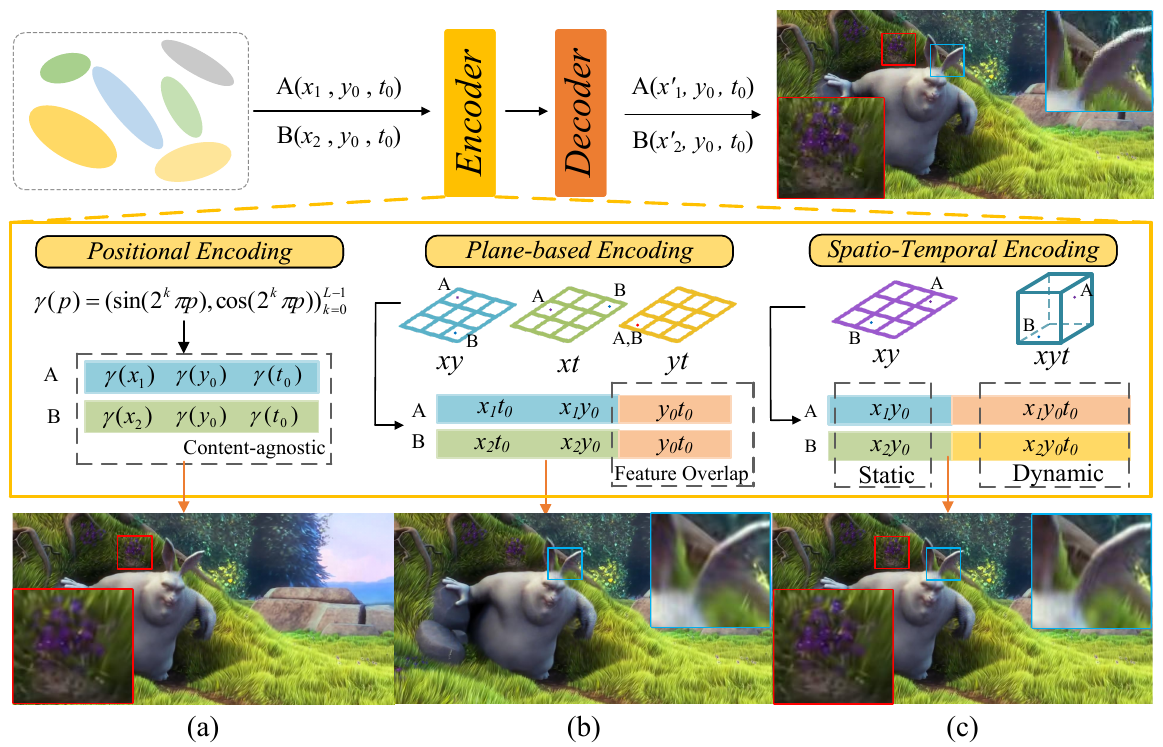}
\end{center}
   \caption{ The visual comparison of utilizing different deformation fields for Gaussian-based video representation. The results show that our method better represent both background details and motion patterns.}
\label{fig:encoding method}
\end{figure}

Motivated by the efficiency and high-fidelity of 3D Gaussian Splatting~\cite{kerbl:hal-04088161_3DGS} in novel view synthesis, recent studies~\cite{Lee_2025_CVPRGaussianvideo, liu2025d2gv, pang2025gsvr} have extended this technique as a 2D Gaussian Splatting (2DGS) framework for video representation by deforming canonical Gaussian primitives across timesteps.
Existing methods typically utilize implicit or explicit deformation fields to deform Gaussian primitives.
However, these methods~\cite{liu2025d2gv},~\cite{Lee_2025_CVPRGaussianvideo} entangle static and dynamic components in videos, hindering the effective learning of their distinct properties and ultimately leading to inaccurate static–dynamic information prediction.
Specifically, implicit deformation field~\cite{liu2025d2gv} (Fig.~\ref{fig:encoding method} (a)) employs positional encodings as embeddings, which are inherently content-agnostic and thus encode static backgrounds and dynamic objects in the same manner.
This simple treatment prevents accurate modeling of the complex appearance and motion differences in videos.
Explicit methods~\cite{Lee_2025_CVPRGaussianvideo, pang2025gsvr} (Fig.~\ref{fig:encoding method} (b)) adopt content-aware feature encodings that decompose spatio-temporal information into three 2D planes.
However, this plane-based decomposition induces spatio-temporal feature collisions for Gaussian primitives with overlapping coordinates, which further entangle static and dynamic information and result in sub-optimal representation performance.

To address these issues, this paper presents a Spatio-Temporal hash encoding framework for high quality Gaussian-based Video representation (STGV), as shown in Fig.~\ref{fig:encoding method} (c).
STGV consists of a 2D spatial and a 3D temporal learnable hash encoding for canonical Gaussian primitive deformations.
Notably, to decouple static and dynamic information in videos, our encoder generates two types of features: spatial features that maintain background details across the timeline, and temporal features that capture motion patterns for dynamic elements.
This substantially reduces spatio-temporal feature overlapping and provides accurate static-dynamic information prediction.
In addition, due to object motion and camera movement in videos, existing methods that utilize multi-frame supervision to initialize the canonical Gaussian primitives tend to be unstable.
These methods average the spatially inconsistent frames into a canonical representation, which induces structurally incoherent geometry and compromises subsequent deformation learning.
Therefore, we present a key frame canonical initialization (KFCI) strategy, which provides a stable and spatially consistent representation for canonical Gaussian primitives.
The experimental results demonstrate that STGV significantly outperforms other Gaussian-based methods in both static and dynamic representation quality.

In general, our contributions can be summarized as the following:
\begin{itemize}[itemsep=0pt, topsep=2pt, leftmargin=1.2em]




    \item We propose a STGV framework that factorizes video features into a 2D spatial and a 3D temporal hash encoding, which alleviates spatio-temporal feature overlapping while disentangling static content and time-varying elements.
    
    \item We introduce a key frame canonical initialization strategy for initializing Gaussian primitives, providing a more stable and spatially consistent representation for subsequent deformation optimization.

    \item Extensive experiments validate the effectiveness of our STGV framework on three widely used video datasets and various downstream tasks, with rapid training speed and real-time decoding up to 625 FPS.
\end{itemize}

\section{Method} 
\label{sec:method}
In this section, we first provide a brief review of 2DGS and our pipeline in Sec.~\ref{subsec:pre_2dgs}. 
Subsequently, we introduce spatio-temporal hash encoding (Sec.~\ref{subsec:spatio-temporal hash}), which separately encodes static and dynamic information for Gaussian primitive deformations.
Finally, we present a key frame canonical initialization (KFCI) strategy (Sec.~\ref{subsec:KFIC}), which provides a stable canonical space for subsequent deformation learning.

\subsection{Preliminary}
\label{subsec:pre_2dgs}
\textbf{2D Gaussian Splatting:} In this paper, we adopt 2DGS representation~\cite{zhang2024gaussianimage} to rasterize each video frame.
2DGS represents an image using a set of 2D Gaussian primitives.
Each Gaussian primitive $\boldsymbol{G}$ is parameterized by its center position $\mu \in \mathbb{R}^2$, color coefficients $c \in \mathbb{R}^3$, and a covariance matrix $\Sigma \in \mathbb{R}^{2 \times 2}$.
During optimization, the covariance matrix $\Sigma$ is factorized using Cholesky decomposition:
\begin{equation}
\Sigma = LL^\top, \quad L = 
\begin{bmatrix}
l_1 & 0 \\
l_2 & l_3 \\
\end{bmatrix},
\end{equation}
where $L$ is a lower triangular matrix parameterized by three learnable parameters $\boldsymbol{l} = \{l_1, l_2, l_3\}$.

While rendering, the final color of each pixel $\boldsymbol{C}_i$ is computed as an accumulative weighted sum by blending all Gaussians that spatially influence the pixel:
\begin{equation}
\boldsymbol{C}_i = \sum_{n \in \mathcal{N}} \boldsymbol{c}_n \cdot \exp(-\boldsymbol{\sigma}_n),
\end{equation}
where $\boldsymbol{\sigma_n}\!=\!\frac{1}{2} \boldsymbol{d}_n^T \Sigma^{-1} \boldsymbol{d}_n$, and $\boldsymbol{d}_n$ measures the displacement between the pixel location and the Gaussian center.

\begin{figure*}[t]
  \centering
  \includegraphics[width=1.0\linewidth]{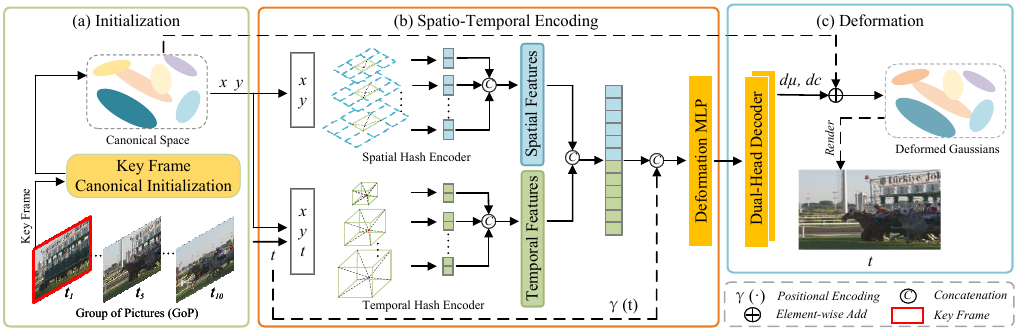}
  \caption{Overall of our proposed method STGV. We first select the first frame from a GoP as the key-frame to perform coarse training to construct the canonical space. Then, the canonical space is trained jointly with the spatio-temporal hash encoding to obtain the deformed attributes of Gaussians at each frame. Finally, the Gaussian attributes of canonical space and the frame-specific Gaussian deformation are combined and rendered to generate the corresponding frame.}
  \label{fig:main_framework}
\end{figure*}

\textbf{Pipeline:} As shown in Fig.~\ref{fig:main_framework}, we split videos into multiple groups of Pictures (GoPs) and train a separate model per GoP, which makes training scalable and alleviates long-range temporal coupling.
For each GoP, we adopt a two-stage training: a coarse stage that initializes a canonical space via a KFCI strategy, and a deformation stage that employs spatio-temporal hash encoding features and a tiny MLP to predict the Gaussian deformations for each corresponding frame.

\begin{figure}[t]
\begin{center}
\includegraphics[width=1.0\linewidth]{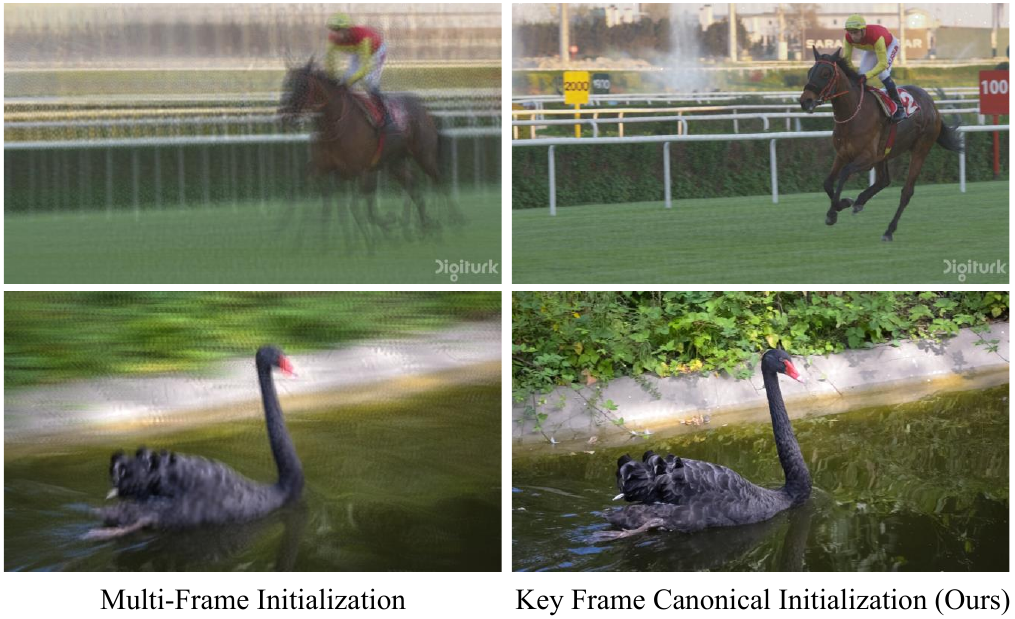}
\end{center}
   \caption{The visual comparison between Key Frame Canonical and Multi-frame Initialization upon canonical space.}
\label{fig:canonical_comparison}
\end{figure}  
\subsection{Spatio-Temporal Hash Encoding}
\label{subsec:spatio-temporal hash}
Videos typically contain static background and dynamic motion that exhibit distinct properties.
Our key insight is that learning a disentangled representation of these components enables more accurate video modeling.
Inspired by~\cite{InstanNGP, Grid4D}, our framework employs a 2D spatial and a 3D temporal hash encodings to decouple the static and dynamic information for deformation learning, as shown in Fig.~\ref{fig:main_framework} (b).

\textbf{Multiresolution Hash Encoder.}
Following~\cite{InstanNGP}, for spatial and temporal hash encoders, we set the multiple resolutions of each dimension in a geometric progression between the coarsest and finest resolutions $[N_{min}, N_{max}]$:
\begin{equation}
N_l = \left\lfloor N_{min} \cdot b^l \right\rfloor, \quad b = \exp\left(\frac{\ln N_{max} - \ln N_{min}}{L - 1}\right),
\end{equation}
where $l \in [0, L-1]$ represents the level index, $L$ is the total number of resolution levels, and $N_l$ indicates the resolution of level $l$. 
Given an input coordinate $\boldsymbol{\mathrm{x}} \in \mathbb{R}^d$, we scale it by $N_l$ and compute coordinates of the neighboring grid vertices by rounding down and up along each dimension $\left\lfloor\boldsymbol{\mathrm{x}}_{l}\right\rfloor=\left\lfloor\boldsymbol{\mathrm{x}} \cdot N_{l}\right\rfloor,\left\lceil\boldsymbol{\mathrm{x}}_{l}\right\rceil=\left\lceil\boldsymbol{\mathrm{x}} \cdot N_{l}\right\rceil.$
We then map each $d$-dimensional integer vertex coordinate $\boldsymbol{\mathrm{x}_l}$ to an entry in the level-$l$ hash table via:
\begin{equation}
h_l(\boldsymbol{\mathrm{x}}_l) = \left( \bigoplus_{i=1,x_i\in\boldsymbol{\mathrm{x}}_l}^{d} x_i \cdot \pi_i \right) \mod T_l ,
\end{equation}
where $\pi_i$ are unique large primes, $\oplus$ denotes bit-wise XOR operation, and $T_l$ is the hash table size at level $l$ that stores learnable feature vectors with dimension $F$.
The encoded features at $\boldsymbol{\mathrm{x}}$ are then computed by multilinear interpolation of the retrieved vertex features and concatenated across all levels:
\begin{equation}
\boldsymbol{f} = Concat(f_0, f_1, \ldots, f_{L-1}) \in \mathbb{R}^{L \cdot F}.
\end{equation}
In our method, given the input coordinate $\mu\!=\!(x, y)$ and $\mu_d\!=\!(x,y,t)$ at time $t$, the spatial and temporal features are obtained from the 2D hash encoder $\boldsymbol{G_{2D}}$ and the 3D hash encoder $\boldsymbol{G_{3D}}$:
\begin{equation}
    \boldsymbol{f}_s = \boldsymbol{G_{2D}}(\mu), \ \ \boldsymbol{f}_d = \boldsymbol{G_{3D}}(\mu_d).
\end{equation}

\textbf{Dual-head Deformation Decoder.}
As illustrated in Fig.~\ref{fig:main_framework} (c), after obtaining the features from the $\boldsymbol{G_{2D}}$ and the $\boldsymbol{G_{3D}}$, we concatenate them with the positional encoding of timestamp $\gamma(t)$ to form the final representation: 
\begin{equation} 
    \boldsymbol{f}_{h} =  Concat(\boldsymbol{f}_s , \boldsymbol{f}_d, \gamma(t)).
\end{equation}
Followed by a two layer MLP $\Phi$ and a dual-head deformation decoder $\mathcal{D} =(\phi_\mu, \phi_c)$, we obtain the deformation of positions and colors:
\begin{equation}
    (\Delta \mu, \Delta c) = (\phi_\mu(\Phi(\boldsymbol{f}_h)), \phi_c(\Phi(\boldsymbol{f}_h)).
\end{equation}
Finally, the deformed attributes of Gaussians are represented as:
\begin{equation}
    (\mu', c', \Sigma) = (\mu + \Delta \mu, c+\Delta c, \Sigma).
\end{equation}
which can be rendered for the corresponding frame at time $t$.

\begin{figure*}[htpb]
  \centering
  \includegraphics[width=1.0\linewidth]{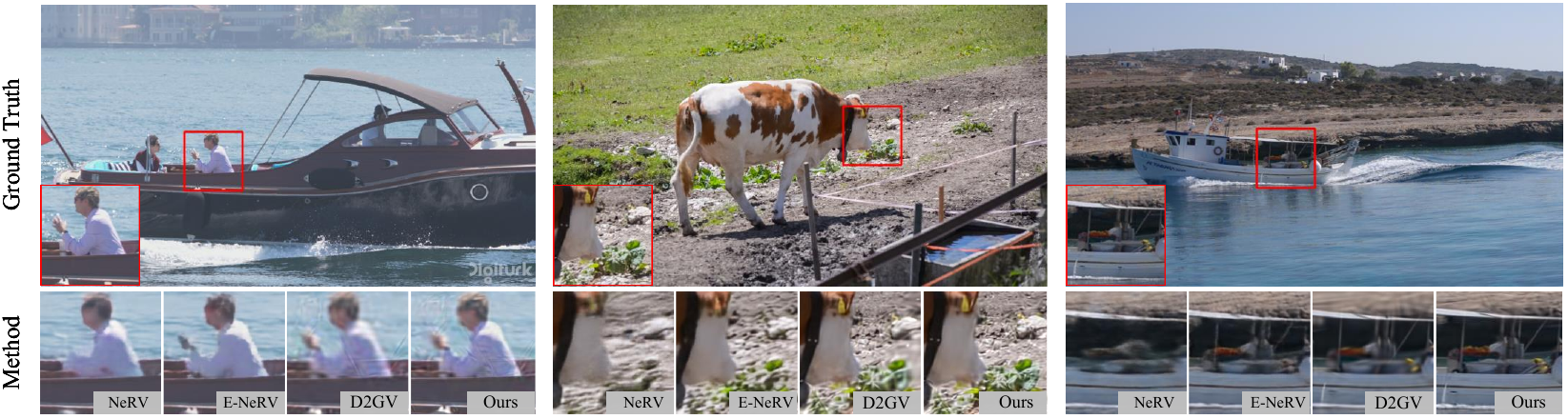}
  \caption{Visual comparison results. From left to right are yachtride, cows and boat video. Benefiting from Spatio-Temporal hash encoding and KFCI strategy, our methods achieves best visual quality.}
  \label{fig:visualizaion comparsion}
\end{figure*}

\subsection{Key Frame Canonical Initialization}
\label{subsec:KFIC}

Video dynamics typically consist of global camera motion and local object motion. 
Under camera movement, the static background remains spatially consistent and can be approximately recovered between frames via simple affine transformations. 
As shown in Fig.~\ref{fig:canonical_comparison}, existing methods initialize canonical Gaussian primitives using multi-frame supervision, which induces blurred textures in the same static regions across frames, since these regions are averaged into a single canonical representation.
Then, the deformation field is forced to explain the inconsistency across frames with a smooth representation rather than modeling the actual dynamic motion, which limits its ability to reconstruct fine details, leading to a structurally incoherent geometric representation and compromising subsequent deformation modeling.

To mitigate this problem, we propose a key frame canonical initialization (KFCI) strategy (Fig.~\ref{fig:main_framework} (a)) that provides a stable and spatially consistent canonical space.
Specifically, we select the first frame of each GoP as the key frame, since it reflects the actual scene content and can serve as a reference for neighboring frames.
We first perform a short coarse stage on the key frame using standard 2DGS, obtaining a canonical space with consistent structure and rich details, and avoiding the blurred averaging effect of multi-frame initialization.
Based on that, the deformation field can focus on modeling the actual dynamic motion while maintaining texture details in the static background, instead of relying on a smoothed representation to compensate for blurred regions, thereby producing more coherent geometry and higher-quality reconstructions.

\section{Experiments}
\label{sec:experiments}
\noindent\textbf{Dataset:} We evaluate our method on the Bunny, UVG~\cite{uvg}, and DAVIS~\cite{Davis} datasets. 
We selected six video sequences from DAVIS.
In our experiment, Bunny is center‑cropped and resized to 640×1280, while DAVIS and UVG are resized to 720p, and we apply 4× framerate subsampling to UVG.

\noindent\textbf{Baselines:} We compare our method against both INR-based and Gaussian-based approaches. For INR methods, we include SIREN~\cite{SIREN}, NeRV~\cite{chen2021nerv}, E-NeRV~\cite{li2022Enerv}, and HNeRV~\cite{chen2023hnerv}.
For Gaussian-based methods, we compare with 2DGS~\cite{zhang2024gaussianimage} trained independently per frame, and its extension D2GV~\cite{liu2025d2gv}, GaussianVideo~\cite{Lee_2025_CVPRGaussianvideo} and GSVR~\cite{pang2025gsvr}.
The model size is measured by the total number of parameters, assuming each parameter is stored in FP32 unless stated otherwise. 

\noindent\textbf{Implementation Details.} We divide videos into GoPs of 10 frames and train separate models for each GoP. 
The number of Gaussians is set to 20K for Bunny and 40K for UVG and DAVIS.
The hidden width of the deformation MLP is 128, with positional encoding frequency of 6 for the timestamp, and optimization uses an $L_2$ loss.
Each group is trained in two stages (10K coarse and 60K deformed steps) on a single NVIDIA 3090 GPU. 
\subsection{Overall Comparison}
\label{subsec:Overallcompare}
We present an overall comparison on the UVG, DAVIS and Bunny datasets, as shown in Table.~\ref{tab:quan_uvg_davis} and Table.~\ref{tab:quan_bunny}. 
Our method achieves superior reconstruction quality, improving PSNR by 0.98 dB and 0.27 dB over Gaussian-based methods, and by 0.20 dB and 2.65 dB over INR-based methods on the UVG and DAVIS datasets, respectively, while maintaining fast training under similar parameter budgets.
In addition, our method attains the second fastest inference speed across all datasets.
The fastest decoding speed of 2DGS on the UVG and DAVIS datasets mainly stems from training each frame of the video independently without deformation modeling.
As for the Bunny dataset, our method achieves performance comparable to HNeRV and surpasses Gaussian-based methods.
These results demonstrate the effectiveness of our design in capturing both static spatial cues and dynamic temporal variations.
Overall, our method achieves better reconstruction quality than Gaussian-based methods and maintains fast inference speed.

We also present the visual comparison of video quality in Fig.~\ref{fig:visualizaion comparsion}.
Compared with other methods, which tend to exhibit blurry textures and ghosting artifacts under motion, our method maintains sharper edges and clearer background textures. More detailed results can be found in the \textbf{Appendix}.
\begin{table}[t]
\centering
\caption{Quantitative results comparison on UVG and DAVIS dataset. 
\textbf{Bold} indicates the best performance, and \underline{underline} indicates the second best.}
\label{tab:quan_uvg_davis}
\resizebox{\columnwidth}{!}{%
\begin{tabular}{l|cccccc}

\multicolumn{6}{c}{\textbf{UVG dataset}} \\
\toprule
\textbf{Model} & \textbf{Training time} & \textbf{Decoding FPS} & \textbf{PSNR} & \textbf{MS-SSIM} & \textbf{Params} \\
\midrule
SIREN           & 8h   & 28              & 26.27                & 0.9197        & \multirow{6}{*}{\centering 5.8M} \\
NeRV            & 5h   & 80              & 33.40                & 0.9494        &                    \\
E-NeRV          & 4h   & 46              & \underline{35.09}    & \textbf{0.9712} &                  \\
2DGS            & 4.5h & \textbf{1800}   & 31.98                & 0.9537        &                    \\
D2GV            & \underline{2.2h} & 310             & 34.31                & \underline{0.9635} &              \\
STGV (Ours)     & \textbf{1.7h} & \underline{460} & \textbf{35.29}       & 0.9600        &                    \\
\midrule
\multicolumn{6}{c}{\textbf{DAVIS dataset}} \\
\toprule
\textbf{Model} & \textbf{Training time} & \textbf{Decoding FPS} & \textbf{PSNR} & \textbf{MS-SSIM} & \textbf{Params} \\
\midrule
SIREN           & 5.8h & 34              & 21.58                & 0.8121        & \multirow{6}{*}{\centering 3.3M} \\
NeRV            & 4.3h & 85              & 26.92                & 0.9013        &                    \\
E-NeRV          & 2.5h & 62              & 28.71                & 0.9288        &                    \\
2DGS            & 4h   & \textbf{1902}   & 30.16                & 0.9491        &                    \\
D2GV            & \underline{1.5h} & 308             & \underline{31.09}    & \underline{0.9611} &              \\
STGV (Ours)     & \textbf{1.3h} & \underline{433} & \textbf{31.36}       & \textbf{0.9613} &                 \\
\bottomrule
\end{tabular}
}
\end{table}

\begin{table}[t]
\centering
\caption{Quantitative results comparison on bunny dataset. 
\textbf{Bold} indicates the best performance, and \underline{underline} indicates the second best.}
\label{tab:quan_bunny}
\resizebox{\columnwidth}{!}{%
\begin{tabular}{l|ccccc}
\toprule
\textbf{Model} & \textbf{Decoding FPS} & \textbf{PSNR} & \textbf{MS-SSIM} & \textbf{Params} \\
\midrule
SIREN           & 29           & 26.27              & 0.9197              & \multirow{8}{*}{\centering 3.0M} \\
NeRV            & 323          & 34.2               & 0.9770              &                                    \\
E-NeRV          & 63           & 37.06              & \underline{0.9887}  &                                    \\
HNeRV           & 96           & \textbf{38.00}     & 0.9880              &                                    \\
GaussianVideo   & 475          & 34.8               & 0.9790              &                                    \\
GSVR            & \textbf{816} & 35.65              & -                   &                                    \\
D2GV            & 403          & 37.26              & 0.9880              &                                    \\
STGV (Ours)      & \underline{625} & \underline{37.93} & \textbf{0.9900}  &                                    \\
\bottomrule
\end{tabular}
}
\end{table}

\subsection{Video compression}
\label{subsec:compression}

Following the compression pipeline of GaussianImage~\cite{zhang2024gaussianimage}, we compress the canonical Gaussian space of each GoP via attribute quantization–aware fine-tuning.
In addition, we apply 8-bit quantization to the parameters of our deformation field to further reduce the overall bit rate.

\begin{figure}[t]
    \centering
    \includegraphics[width=\columnwidth]{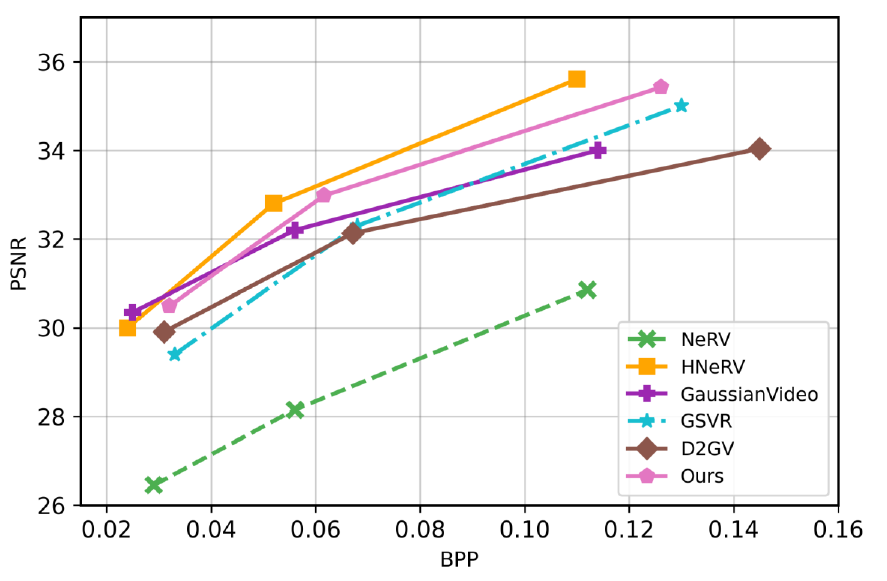}
    \caption{Compression results on Bunny dataset.}
    \label{fig:bunny_rd}
\end{figure}

As shown in Fig.~\ref{fig:bunny_rd}, our rate–distortion performance is superior to other Gaussian-based methods, indicating a more favorable rate–distortion trade-off.

\subsection{Downstream Tasks}
\label{subsec:downstream}
\paragraph{Video Inpainting} 
\label{para:video inpating}
We evaluate video restoration performance using five videos from UVG and DAVIS. 
For each video, we randomly apply five $50 \times 50$ masks per frame and ignore the masked regions during training loss computation.   

\begin{table}[b]  
\centering
\label{tab:Inpainting}
\caption{PSNR results for video inpainting. "baseline" means the mask frames before inpainting}  
\resizebox{\columnwidth}{!}{%
\begin{tabular}{l|ccccc|c}
\toprule
\textbf{Method} & \textbf{Bee} & \textbf{Swan} & \textbf{Cows} & \textbf{Bos.} & \textbf{Beauty} & \textbf{Avg.} \\
\midrule
Baseline     & 27.08 & 26.25 & 24.47 & 22.67 & 29.13 & 25.92  \\
D2GV         & 40.32 & 29.89 & 29.30 & 33.91 & 35.49 & 33.78  \\
STGV      & 41.85 & 30.21 & 30.05 & 34.01 & 36.41 & 34.51  \\
\bottomrule
\end{tabular}
}
\end{table}
\begin{figure}[b]
\begin{center}
\includegraphics[width=1.0\linewidth]{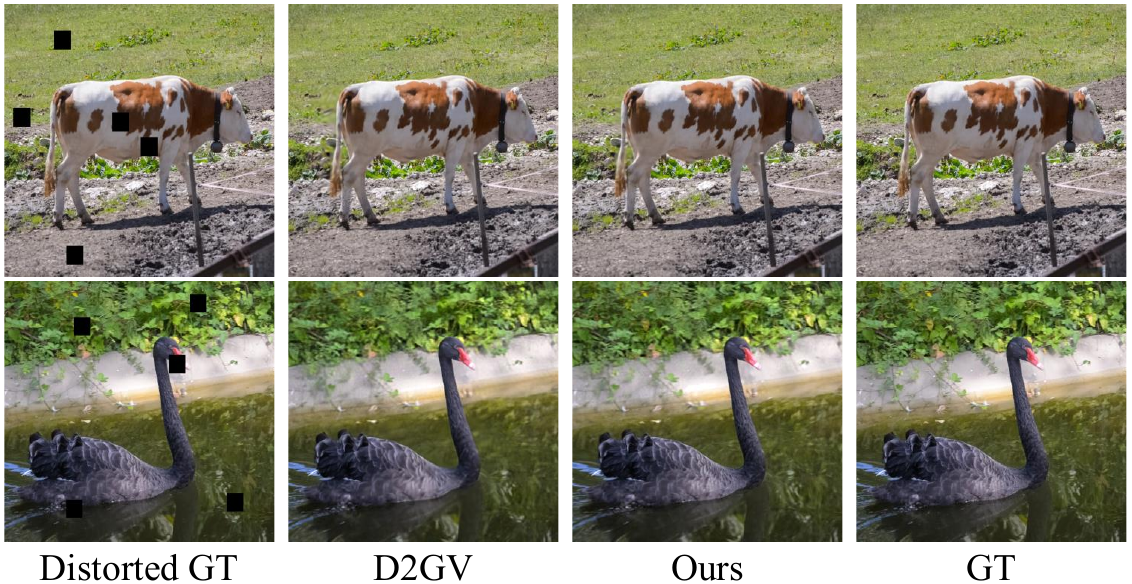}
\end{center}
   \caption{Video inpainting visualization on cows and blackswan.}
\label{fig:inpainting}
\end{figure}
As shown in Fig.~\ref{fig:inpainting}, our method successfully restores occluded regions without introducing artifacts or noise. 
While NeRV employs implicit modeling to maintain temporal continuity, and D2GV ignores explicitly modeling spatial Gaussian relationships, our method benefits from spatio-temporal hash-encoding to capture fine details. 

\begin{figure}[t]
\begin{center}
\includegraphics[width=1.0\linewidth]{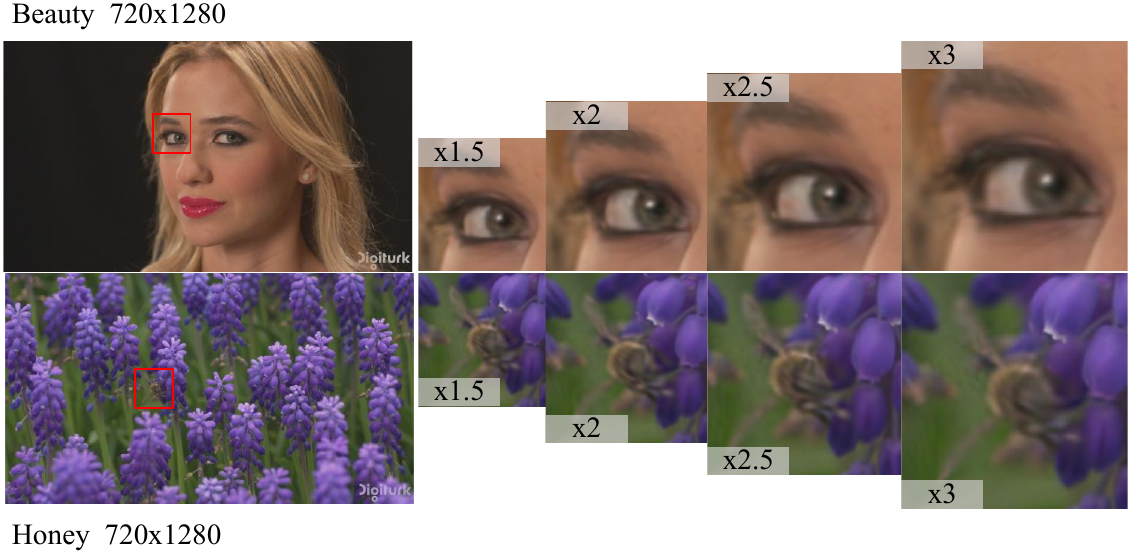}
\end{center}
   \caption{The spatial interpolation visualization on Beauty and Honeybee.}
\label{fig:spatial interpolation}
\end{figure}

\paragraph{Spatial Interpolation}
Based on the continuous nature of 2D Gaussian splatting, we directly render the Gaussians at higher resolutions and obtain frames at arbitrary scales without retraining.
We test scale factors from x1.5 to ×3.0 on Honeybee and Beauty. 
As shown in Fig.~\ref{fig:spatial interpolation}, our approach preserves fine details and sharp structures, benefiting from explicit Gaussian primitives and our spatio-temporal deformation field, thereby showing strong potential for video super-resolution.

\begin{table}[b]
\centering
\caption{Ablation study on Bunny and UVG datasets in PSNR. \textbf{KFCI} means Key Frame Canonical Initialization strategy, \textbf{Sta.} represents static hash encoding and \textbf{Dyn.} means dynamic hash encoding.}
\label{tab:ablation_UVG}
\begin{tabular*}{\columnwidth}{@{\extracolsep{\fill}} l|ccccc}
\toprule
\textbf{Variants} & \textbf{KFCI} & \textbf{Sta.} & \textbf{Dyn.} & \textbf{Bunny} & \textbf{UVG} \\
\midrule
(V1) &           &             &            & 37.26 & 34.93 \\
(V2) & \checkmark &            &            & 37.81 & 35.01 \\
(V3) & \checkmark & \checkmark &            & 37.64 & 35.19 \\
(V4) & \checkmark &            & \checkmark & 37.33 & 35.10 \\
Ours & \checkmark & \checkmark & \checkmark & 37.93 & 35.29 \\
\bottomrule
\end{tabular*}
\end{table}
\subsection{Ablation Study}
\label{subsec:aba}

To validate the effectiveness of each component in our method, we conduct ablation experiments on the Bunny and UVG datasets. 
The results are summarized in Table.~\ref{tab:ablation_UVG}.

Without the hash encoding and KFCI strategy, the variant (V1) is based on positional encoding.
Compared to (V1), (V2) improves the reconstruction quality by PSNR 0.55 dB and 0.08 dB on the Bunny and UVG datasets, providing a stable and spatially consistent representation with the KFCI strategy.  
As the Bunny dataset exhibits weak camera movement and the motion of the big rabbit, our KFCI strategy allows the deformation field to learn the actual dynamic motion while maintaining texture details in static background regions, rather than relying on a smoothed representation to compensate for blurred regions.

\begin{figure}[t]
\begin{center}
\includegraphics[width=1.0\linewidth]{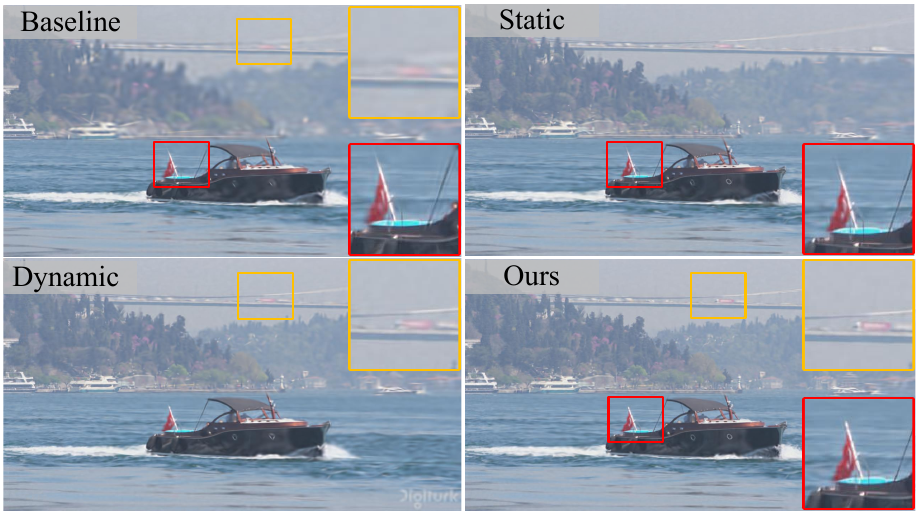}
\end{center}
   \caption{Visual comparison of static hash encoding and dynamic encoding in bosphorus dataset. Our method achieves the best quality in the dynamic region (red block) and the static region (yellow block).}
\label{fig:Static_dynamic}
\end{figure}

Next, we evaluate the impact of static hash (V3) and dynamic hash (V4) encoding. 
Compared with (V2), introducing static or dynamic hash encoding improves the reconstruction quality by 0.18 dB and 0.09 dB in PSNR on the UVG dataset, since they avoid employing content-agnostic embeddings for encoding video features.
Notably, the degraded quality of (V3) and (V4) on the Bunny dataset mainly stems from the ambiguity induced by the combination of significant object motion and subtle camera movement, which leads to severe feature entanglement and blurry reconstruction.
In contrast, our method achieves superior performance to (V2) by PSNR 0.12 dB and 0.28 dB on the Bunny and UVG datasets, since we decompose the static and dynamic information in videos, maintaining background details while effectively capturing motion patterns.
As shown in Fig.~\ref{fig:Static_dynamic}, our method achieves the best visual quality, whereas directly employing only the static and dynamic hash encoding introduces a blurred representation in dynamic and static regions.

Finally, we also conduct experiments with different encoding methods on the Bunny and UVG dataset as shown in Table.~\ref{tab:ablation_Davis}. Our method achieves the best performance against other methods, as our method disentangled static and dynamic feature learning, while avoiding content-agnostic or spatio-temporal feature overlapping embeddings to predict the deformation.

\begin{table}[t]
\centering
\caption{Different encoding methods results on Bunny and UVG datasets in PSNR.}
\label{tab:ablation_Davis}
\begin{tabular*}{\columnwidth}{@{\extracolsep{\fill}} l|ccc}
\toprule
 & \multicolumn{3}{c}{\textbf{Encoding method}} \\ 
\cmidrule{2-4} 
\textbf{Dataset} & \textbf{Positional} & \textbf{Plane-based } & \textbf{Spatio-Temporal} \\
\midrule
Bunny   & 37.81 & 37.55 & 37.93\\
UVG     & 35.01 & 34.00 & 35.29 \\
\bottomrule
\end{tabular*}
\end{table}


\section{Conclusion}
\label{sec:conclue}
In this paper, we propose a novel 
framework for efficient video representation via deformable 2D Gaussian Splatting.
Our spatio-temporal hash encoding enables disentangled static and dynamic feature learning for accurate deformation. 
Furthermore, we introduce a key frame canonical initialization strategy to provide a stable and spatially consistent canonical space for deformation learning.
Extensive experiments demonstrate that our method achieves competitive reconstruction quality while offering significant advantages in training time and decoding speed.

\section{Acknowledgment}
\label{sec:Acknow}
This work was supported by the National Natural Science Foundation of China (Grant No. 62571160), Engineering Technology R\&D Center of Guangdong Provincial Universities (2024GCZX004), and the Pengcheng Laboratory.

\bibliographystyle{IEEEbib}
\bibliography{icme2026references}

\clearpage
\appendix

\section{Supplementary Materials}
In this supplementary material, we begin with the review of the related work.
Then, we provide additional details about our implementation.
Next, we offer additional visual comparisons on different dataset and the detailed results on each video.
Then, we analysis the results on video denoise and present the visual comparison. 
The source code can be found at \url{https://github.com/JRL-szu/STGV.git}.

\subsection{Related Work}
\label{subsec:relat}
\textbf{Implicit nerual representation:} Implicit neural representations (INRs) is an innovative approach for modeling visual data by encoding signals such as images or videos within neural networks.  
This paradigm has been successfully applied to novel view synthesis, audio modeling, and image/video representation.

For video tasks, SIREN~\cite{SIREN} employs multi-layer perceptrons (MLPs) to directly map spatio-temporal coordinates to RGB values, achieving continuous and temporally coherent representations.
However, it requires pixel-wise sampling during inference, resulting in significant computational cost.
To improve decoding efficiency, NeRV~\cite{li2022nerv} replaces pixel-wise regression with a convolutional architecture that maps temporal indices to entire video frames, significantly accelerating inference.
E-NeRV~\cite{li2022Enerv} further introduces spatial decoupling to reduce redundancy and enhance representation capacity, while HNeRV~\cite{chen2023hnerv} incorporates an additional encoder for learning frame embeddings as input conditions. 
FFNeRV~\cite{FFneRV} and HiNeRV~\cite{HiNeRV} leverage feature grids to improve latent representation learning. 
DNeRV~\cite{Zhao_2023_CVPR_DNeRV} focuses on large-motion videos by combining spatial features and dynamic information from neighboring frames. 
DS-NeRV~\cite{Yan_2024_CVPR_DSNeRV} disentangles static and dynamic components and uses attention-based fusion to better capture long-range temporal structures and transient motion. 
Tree-NeRV~\cite{Zhao_2025_ICCVTreeneRV} dynamically organizes a binary tree of latent key frames to model hierarchical dependencies across time.

While these methods have demonstrated promising results in compact and high-quality video modeling, INR-based approaches still suffer from high model complexity, slow training and inference, and limited interpretability.

\textbf{2D Gaussian Splatting:} To leverage the explicit and differentiable rasterization pipeline of Gaussian Splatting for image-level tasks, GaussianImage\cite{zhang2024gaussianimage} introduced a novel paradigm for image representation and compression. 
By modeling an image as a set of 2D Gaussians, it enables a compact and expressive representation that supports high-fidelity reconstruction, fast training, and efficient rendering.

Base on this foundation, 2DGS has been applied to various downstream tasks. For example,~\cite{peng2025pixelgaussianultrafastcontinuous} proposes the ContinuousSR framework for image super-resolution, where 2D Gaussians are used to reconstruct continuous high-resolution signals directly from low-resolution inputs.
In 3D scene domain, HybridGS~\cite{Lin_2025_CVPRHybridGS} introduces a hybrid representation that decouples static and transient components by combining 3D and 2D Gaussian Splatting, where static background content is modeled using 3D Gaussians as in standard 3DGS pipelines, while transient objects are represented with 2D Gaussians estimated from a single input view.

More recently, several works have extended 2DGS to video representation by incorporating temporal deformation mechanisms. 
Notably, D2GV~\cite{liu2025d2gvdeformable2dgaussian} constructs a shared canonical Gaussian space for each group of pictures (GoP) and uses a tiny MLP to learn the frame-specific deformation. 
This approach achieves fast decoding, high reconstruction quality with a small model size, and strong adaptability to downstream tasks. 
However, it adopts the content-agnostic positional encodings as embeddings, which limits its capacity for modeling complex variation.
 
Similarly, GaussianVideo~\cite{Lee_2025_CVPRGaussianvideo} establishes a shared canonical space and encodes temporal variation through grid-based feature extraction, followed by an MLP that predicts frame-wise deformations.
While this plane-based decomposition induces spatio-temporal feature collisions for Gaussian primitives with overlapping coordinates, which further mixes static and dynamic information and leads to suboptimal representation performance.

\subsection{Implementation Details}
\label{subsec:imple-detail}
In our implementation, we train separate model for each GoP using a two-stage pipeline.
During the coarse stage, we choose the first frame of the GoP as the key frame and employ the standard 2DGS pipeline to reconstruct it, which will be treat as the initial canonical space in deformation stage.
The training epochs for this stage is set to 10000, which can be completed within half a minute.
In deformation stage, we adopt the spatio-temporal hash module to decompose the video feature into spatial and temporal features.
Followed by a two-layer MLP with a width of 128 and a multi-head decoder to decode the deformation of positions and colors.
We incorporate the timestamp positional encoding to serve as a smooth regularizer, which the frequency is set to 6.
The spatio-temporal hash module consist of a 2D hash encoding and a 3D hash encoding.
As for the 2D hash encoding, it employ 8 resolution levels, where each level stores 2 feature channels in a hashed table of size $2^{10}$.
The base grid resolution is 16, and the resolution increases by a factor of 1.5 per level.
The 3D hash encoding uses the same hyperparameters as the 2D one, except that it operates on 3D inputs and uses 4 features per level.  
The initial learning rate for spatio-temporal hash module and MLP is set $1.6 \times 10^{-4}$, and the final learning rate to $1.6 \times 10^{-5}$.
Both the stage set the learning rate of 2D Gaussian to $7\times10^{-3}$.
During the entire training, we employ the $L_2$ loss between the rendered and ground-truth frames.
The epochs in deformation stage is set to 60000.

\subsection{Video Compression Pipeline}
Following the compression pipeline in ~\cite{zhang2024gaussianimage}, we compress the Gaussian in canonical space of each group through attribute quantization-aware fine-tuning.  
Specifically, we adopt 16-bit float precision for position parameters to preserve reconstruction fidelity.
Then, we incorporate a 8-bit asymmetric quantization technique for cholesky vector $l_n$, where both the scaling factor $\gamma_i$ and the offset factor $\beta_i$ are learned during fine-tuning:
\begin{equation}
\hat{l}_{i}^{n}=\left\lfloor\operatorname{clamp}\left(\frac{l_{i}^{n}-\beta_{i}}{\gamma_{i}}, 0,2^{b}-1\right)\right\rfloor, \bar{l}_{i}^{n}=\hat{l}_{i}^{n} \times \gamma_{i}+\beta_{i}. 
\end{equation}
As for weighted color coefficients, we employ residual vector quantization(RVQ) that cascades $M$ stages of VQ with codebook size $B$ to achieve representative color attribute encoding:
\begin{equation}
\begin{split}
\hat{c}_n^{\prime m} &= \sum_{k=1}^{m} \mathcal{C}^{k}\!\left[i^{k}\right], \quad m \in \{1, \dots, M\}, \\
i_n^m &= \arg\min_{k} \left\| \mathcal{C}^{m}[k] - \left( c_n^{\prime} - \hat{c}_n^{\prime\, m-1} \right) \right\|_2^2, \quad \hat{c}_n^{\prime\, 0} = \mathbf{0}.  
\end{split}
\label{eq:iterative_coding}
\end{equation}
During training the codebooks, we apply the commitment loss $L_c$ as follows:
\begin{equation}
    L_c = \frac{1}{N\times B}\sum_{k=1}^M \sum_{n=1}^N \| \operatorname{sg}[c_n' - \hat{c}_n'^{k-1}] - C^k[i_n^k]\|^2_2.
\end{equation}
where $N$ represent the number of Gaussian and $\operatorname{sg}[\cdot]$ is the stop-gradient operation.

During the compression task, the overall loss $\mathcal{L}$ consists of the $L_2$ loss and the commitment loss $L_c$:
\begin{equation}
    \mathcal{L} = L_2 + \lambda L_c
\end{equation}
where $\lambda$ serves as the hyper-parameter, balancing the weight of each loss component. In our experiment, we set $\lambda$ to 0.1.
The color codebooks are initialized using the K-means algorithm, providing a robust starting point for subsequent optimization. 
During fine-tuning, we adopt the exponential moving average mode to update the codebook.

To reduce the overall bit cost of the deformation field, we apply 8-bit quantization to both the spatio-temporal hash module and the MLP. In our experiments, this quantization has a negligible impact on reconstruction quality.

\subsection{Video Denoising}
We evaluate the denoising capability of STGV on the HoneyBee video as shown in Table.~\ref{tab:denoising}, corrupted with three representative noise types: additive white Gaussian noise, uniform black noise, and salt-and-pepper noise. 
We compare our method with D2GV and classical denoising techniques, including Gaussian filtering, uniform spatial averaging, median filtering, morphological erosion (min-pooling), and dilation (max-pooling), all implemented with OpenCV.

The results show that STGV achieves strong denoising performance without requiring any explicit noise suppression module. 
Its reconstruction quality is comparable to models trained on clean data, showcasing its robustness to noise. However, due to the high-frequency preference inherent in hash encoding, STGV may risk overfitting to noise patterns. 
In future work, incorporating smoothness constraints in the deformation field may help improve its denoising generalization and achieve more balanced results.

\begin{table}[t]
\centering
\caption{Video denoising performance. PSNR is recorded in dB.}
\label{tab:denoising}
\resizebox{\columnwidth}{!}{%
\begin{tabular}{lcccccc}
\toprule
\textbf{Noise} & \textbf{White} & \textbf{Black} & \textbf{Salt \& Pepper} & \textbf{Avg} \\
\midrule
\textbf{Baseline}     & 26.94	&	28.18	&	28.35	&	27.82	\\
Gaussian     &	35.64	&	36.08	&	36.46	&	36.06  \\
Uniform      &	34.35 	&	34.35	&	34.60	&	34.43  \\
Median       &	34.73 	&	38.07	&	38.15	&	36.98  \\
Minimum      &	18.64 	&	14.3	&	18.75	&	18.75  \\
Maximum      &	18.86 	&	23.4	&	15.15	&	19.14  \\
\midrule
D2GV         & 40.14 & 39.44 & 40.50 & 40.03  \\
\textbf{Ours}      & 40.19 & 39.60 & 40.38 & 40.06  \\
\bottomrule
\end{tabular}
}
\end{table}

\begin{figure}[b]
    \centering
    \includegraphics[width=\columnwidth]{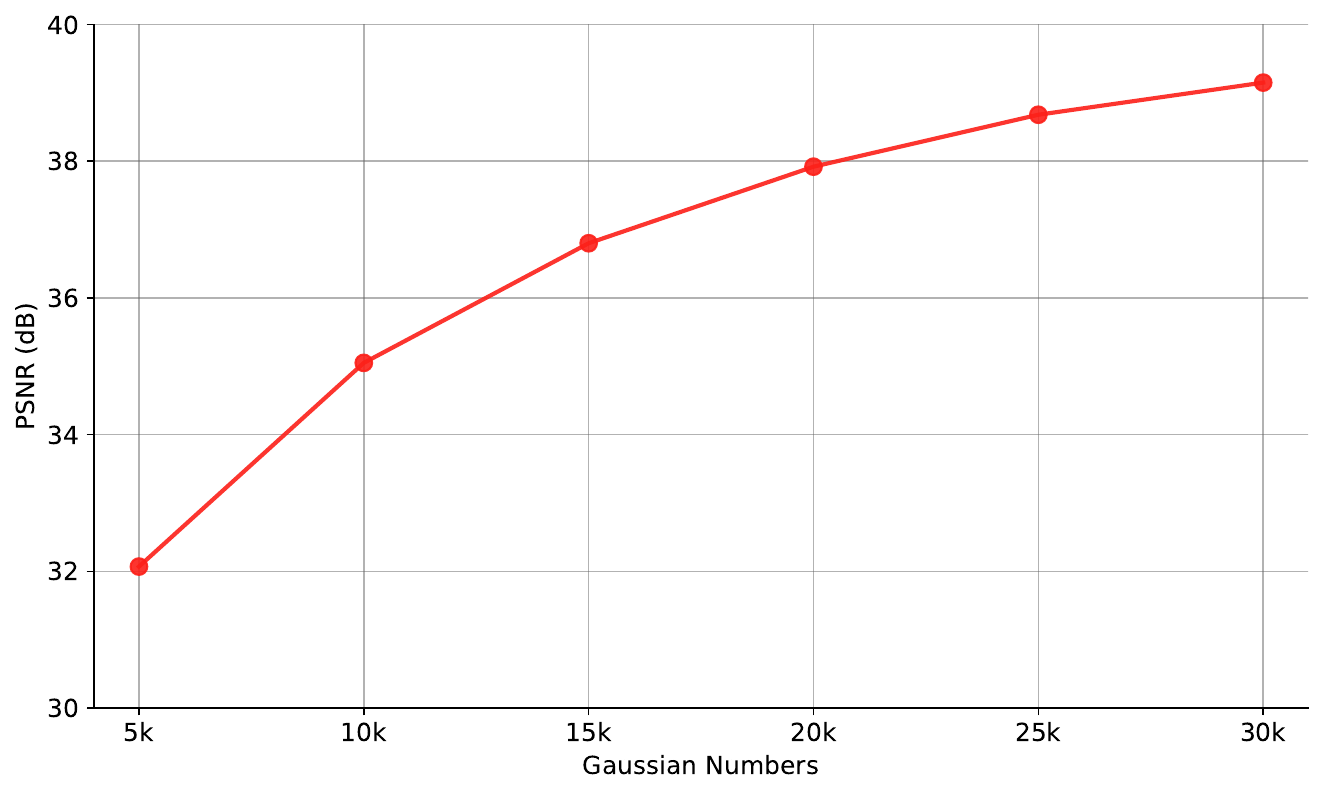}
    \caption{Results with different Gaussian Numbers.}
    \label{fig:Diff_GSn}
\end{figure}

\begin{figure*}[t]
  \centering
  \includegraphics[width=1.0\linewidth]{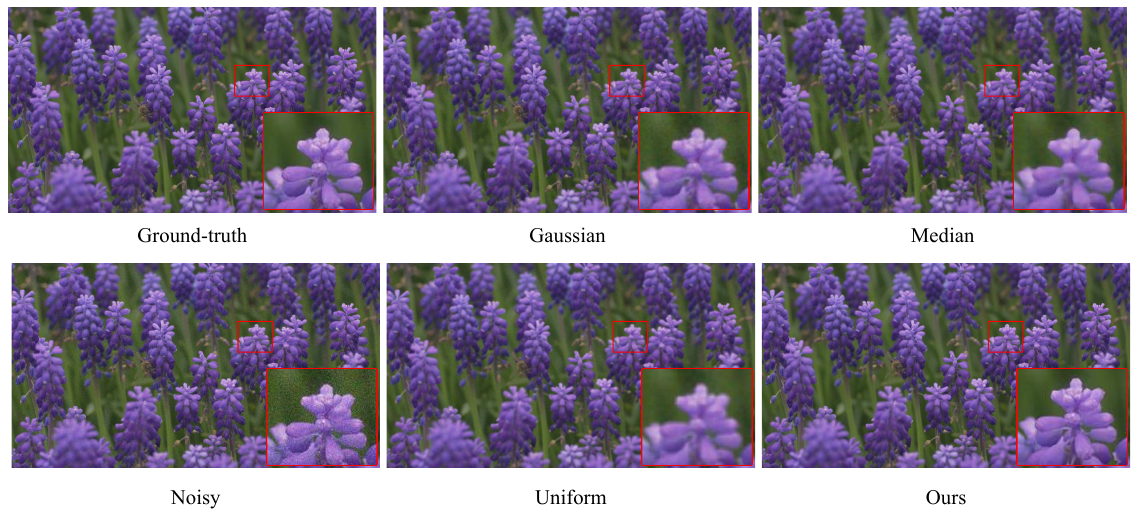}
  \caption{Visualization of video denoising.}
  \label{fig:Denoise_visual}
\end{figure*}
\begin{table*}[t]
\centering
\caption{Quantitative results comparison on the UVG dataset in PSNR/MS-SSIM.}
\label{tab:quan_each_uvg}
\resizebox{\textwidth}{!}{%
\begin{tabular}{l|ccccccc|c}
\toprule
Method & Beauty & Bosph. & Honey. & Jockey & Ready. & Shake. & Yacht. & Avg. \\
\midrule
NeRV & 35.24/0.9446 & 33.95/0.9567 & 39.88/0.9924 & 34.07/0.9509 & 27.00/0.9324 & 34.98/0.9667 & 28.67/0.9212 & 33.40/0.9494 \\
E-NeRV & 35.82/0.9508 & 36.01/0.9760 & 39.06/0.9937 & 36.09/0.9710 & 30.30/0.9689 & 36.53/0.9790 & 31.34/0.9590 & 35.09/0.9712 \\
2DGS & 33.71/0.9419 & 33.09/0.9599 & 32.42/0.9921 & 34.01/0.9540 & 28.10/0.9410 & 30.12/0.9460 & 31.34/0.9407 & 31.98/0.9537 \\
D2GV & 35.89/0.9510 & 34.42/0.9705 & 40.62 / 0.9940 & 35.54 / 0.9688 & 28.97 / 0.9380 & 35.91 / 0.9792 & 29.79 / 0.9428 & 34.31/0.9635 \\
Ours & 36.68/0.9571 & 34.31/0.9505 & 41.87/0.9943 & 36.04/0.9597 & 29.63/0.9400 & 37.13/0.9764 & 31.38/0.9420 & 35.29/0.9600 \\
\bottomrule
\end{tabular}
}
\end{table*}
\begin{table*}[t]
\centering
\caption{Quantitative results comparison on the Davis dataset in PSNR/MS-SSIM.}
\label{tab:quan_each_davis}
\resizebox{\textwidth}{!}{%
\begin{tabular}{l|cccccc|c}
\toprule
Method & Dance & Camel & Bmx & Swan & Cow & Boat & Avg. \\
\midrule
NeRV & 28.10/0.9678 & 24.85/0.8860 & 27.04/0.8897 & 28.12/0.9061 & 22.74/0.8099 & 31.69/0.9482 & 26.92/0.9013 \\
E-NeRV & 30.28/0.9861 & 26.58/0.9293 & 29.27/0.9325 & 30.51/0.9502 & 22.65/0.8102 & 32.94/0.9645 & 28.71/0.9288 \\
2DGS & 30.51/0.9801 & 29.71/0.9431 & 29.01/0.9221 & 30.27/0.9299 & 29.01/0.9641 & 33.01/0.9554 & 30.16/0.9491 \\ 
D2GV & 31.38/0.9835 & 30.89/0.9734 & 30.78/0.9340 & 30.42/0.9371 & 29.79/0.9700 & 34.31/0.9680 & 30.26/0.9611 \\
Ours & 31.14/0.9814 & 30.65/0.9630 & 30.44/0.9367 & 30.33/0.9382 & 30.12/0.9731 & 35.47/0.9756 & 31.36/0.9613 \\
\bottomrule
\end{tabular}%
}
\end{table*}
\begin{figure*}[t]
  \centering
  \includegraphics[width=1.0\linewidth]{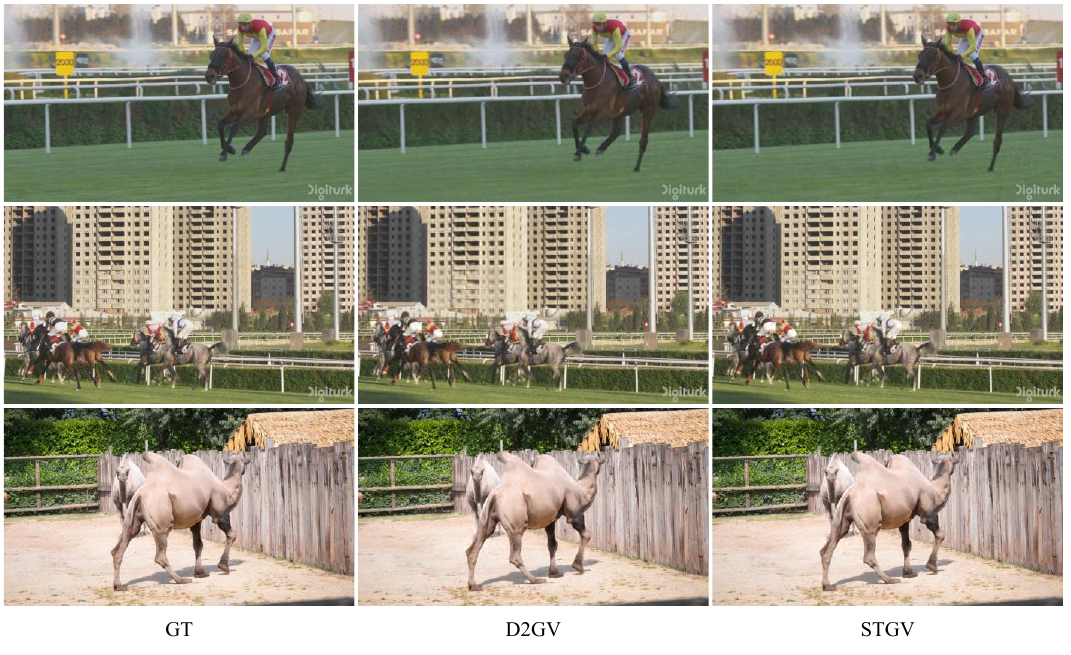}
  \caption{Visualization of reconstruction quality on Jockey, Readysteaygo, and Camel (from top to bottom).}
  \label{fig:visual_appendix}
\end{figure*}

\subsection{Detailed Results}
We present the per video PSNR and MS-SSIM assessment against other baselines in Table.~\ref{tab:quan_each_uvg} and Table.~\ref{tab:quan_each_davis}. 
Our method achieves quality on par with or exceeding stateof-the-art (SOTA) INRs. 
Notably, our approach adopts spatio-temporal hash encoding to capture the local high-frequency details and disentangled the static and dynamic information for deformation learning.
Benifiting from that, our method achieves the best visual quality and the fastest decoding fps.

Further more, we present the results with different Gaussian Numbers in Fig.~\ref{fig:Diff_GSn}, and more visual results on Jockey, Readysteaygo, and Camel from UVG and DAVIS in Fig~\ref{fig:visual_appendix}. We compare our method with D2GV, while our method achieves the better performance.

\subsection{Future Work}
Although the spatio-temporal hash encoding in STGV disentangles static and dynamic information, leading to more precise prediction of Gaussian primitives and improved reconstruction quality, the inherent high-frequency bias of multiresolution hash encoding may still cause overfitting to transient noise. In future work, we plan to incorporate a smoothness regularization loss to further improve the performance of our method.
In addition, STGV selects the first frame of each GoP as the key frame, which may fail to fully reflect the content of other frames within the same GoP. The adaptive selection of key frames remains an open problem and will be further investigated.

\end{document}

\end{document}